# A neural network model of perception and reasoning

Paul J. Blazek[1,2], Milo M. Lin[1,2,3,*]

## Abstract

How perception and reasoning arise from neuronal network activity is poorly understood. This is reflected in the fundamental limitations of connectionist artificial intelligence, typified by deep neural networks trained via gradient-based optimization. Despite success on many tasks, such networks remain unexplainable black boxes incapable of symbolic reasoning and concept generalization. Here we show that a simple set of biologically consistent organizing principles confer these capabilities to neuronal networks. To demonstrate, we implement these principles in a novel machine learning algorithm—based on concept construction instead of optimization—to design deep neural networks that reason with explainable neuron activity. On a range of tasks including NP-hard problems, their reasoning capabilities grant additional cognitive functions, like deliberating through self-analysis, tolerating adversarial attacks, and learning transferable rules from simple examples to solve problems of unencountered complexity. The networks also naturally display properties of biological nervous systems inherently absent in current deep neural networks, including sparsity, modularity, and both distributed and localized firing patterns. Because they do not sacrifice performance, compactness, or training time on standard learning tasks, these networks provide a new black-box-free approach to artificial intelligence. They likewise serve as a quantitative framework to understand the emergence of cognition from neuronal networks.

## Introduction

One of the defining features of living systems is the ability to integrate multiple signals and respond appropriately. However, it is often difficult to understand these processes mechanistically because they involve multiple distributed agents[1]. For human cognition, the ancient question of how the brain gives rise to thought has been embodied during the last century in the divide between symbolic cognitive models and connectionist neural network models[2,3]. On the one hand, symbolic models of reasoning have historically been phenomenological, thereby lacking a direct mechanistic link to the brain's neuronal structure. On the other hand, connectionist models of neural networks have been unable to explain the emergence of conceptual thought and symbolic manipulation. The divide between symbolism and connectionism has been especially evident in their various implementations in artificial intelligence (AI) systems. Recent attempts at capturing symbolic reasoning in connectionist models have not addressed this divide because they are hybrids in which symbolic work only occurs on the outputs of networks, ignoring the biological need to integrate symbolic manipulation within the network itself[4,5].

As a consequence of this foundational divide, both symbolic and connectionist AI currently suffer from fundamental limitations. Symbolic AI often requires problem-specific manual tuning, does not scale well to combinatorially large problems, and is not able to learn features from raw input data[2,6,7]. Connectionist deep learning models have recently become popular due to their superhuman accuracy across a large range of tasks[2,6,7,8,9]. This approach utilizes artificial neural networks with layers of biologically inspired neurons which learn by gradient descent with backpropagation. Despite its success, it is widely regarded as a black box because there currently exists no explanation for the learned synaptic weights or for the process by which individual neurons give rise to the final output[3,10,11]. Deep learning does not generalize concepts and therefore requires training on large labeled datasets from the same distribution as the desired task[3,7,12]. The networks are also paradoxically fooled by adversarial attacks with small, human-imperceptible input perturbations[3,11]. On a more basic level, general features observed in biological neural networks such as modularity, hubs, and sparse neuron firing are not naturally learned by artificial neural networks[13,14]. Despite significant efforts[3,4,7,10,15], these problems remain largely unresolved.

[1] Green Comprehensive Center for Molecular, Computational and Systems Biology, University of Texas Southwestern Medical Center, Dallas, TX, 75390

[2] Department of Biophysics, University of Texas Southwestern Medical Center, Dallas, TX, 75390

[3] Center for Alzheimer's and Neurodegenerative Diseases, University of Texas Southwestern Medical Center, Dallas, TX, 75390

*Correspondence to: milo.lin@utsouthwestern.edu

Here we develop a biologically consistent model of neural networks based on the philosophy of essences that bridges the divide between connectionism and symbolism. We implement it in a novel machine learning algorithm to train what we call essence neural networks (ENNs). We show that ENNs are inherently explainable and capable of hierarchical organization, deliberation, symbolic manipulation, concept generalization, and using both distributed and localized representations. Consequently, ENNs are also scalable, sparse, modular, and significantly more robust to noise and adversarial attacks.

## A neural network model of symbolic categorization

Human reasoning has been studied for millennia. Classical and medieval philosophers such as Aristotle and Thomas Aquinas described how the mind grasps the essence of a particular conceptual species by its definition, which includes a super-categorical genus and a set of qualities, or differentiae, that separate the concept from other intrageneric concepts[16,17,18,19,20]. Perception and reasoning have been described similarly as a series of divisions and compositions[21,22,23,24,25]. The genus-differentia model is also consistent with the modern decision-boundary theory of perception, as distinct from prototype and exemplar cognitive models[26,27].

However, given a collection of input features, in many cases no set of differentiae exists that adequately define all members of a conceptual species[28]. The definition of a concept may therefore require division into a plurality of subconcepts. For example, subconceptualization could be necessary to recognize animals with life-cycle morphological changes, to understand words with multiple meanings, or to visually distinguish fruits and vegetables by first identifying the specific type of produce (Fig. 1).

With this schema we have developed a new model called essence neural networks (ENNs). In ENNs, (i) a set of inputs are observed and (ii) compared to all of the network's differentiae, the results of which are (iii) unified to determine a subconcept, and then (iv) assigned to an output concept or decision (Fig. 1). While this approach can be generalized to multiple rounds of divisions and compositions, we have mostly focused on this simple four-layer architecture.

To map this process to individual neurons, we used the insight from conceptual space theory that natural concepts exist as convex sets in the space of relevant features[6,29]. The linear separability of disjoint convex sets is useful here because artificial neurons are typically modeled mathematically as hyperplanes[30]. The output of a neuron is given by $\sigma(\boldsymbol{w} \cdot \boldsymbol{x} + b)$, which is a function of the distance of the incoming signal $x$ from the neuron's hyperplane $w$ with bias $b$, and the sigmoid activation function $\sigma(x) = \frac{1}{1+e^{-x}}$ saturates the output between non-firing (0) and maximal firing (1). It therefore seems natural to model each neuron as responsible for separating, or distinguishing, concepts.

We developed a novel machine learning algorithm to train ENNs (Appendix A), which first clusters training samples into subconcepts within each conceptual class to form more easily separable convex sets. Differentia neurons are generated by computing linear support-vector machines (SVMs) between pairs of subconcepts, the outputs thereby measuring the relative similarity to one subconcept versus another. These differentia outputs are used to generate subconcept neurons, again using SVMs but now separating each subconcept from all other inter-class subconcepts to measure similarity to each subconcept. During this stage, less important differentiae are pruned from the network. Finally, subconcept outputs are used to generate the concept neurons, using SVMs to separate each concept from all others, optionally adjusting these final weights so that the outputs yield useful probabilities. ENN training therefore is inherently a category-based process that learns how concepts are structured and defined by input features. This is fundamentally different from the current optimization-based approach to deep learning, which focuses solely on mapping inputs to outputs by minimizing a loss function over the whole network, using stochastic gradient descent and backpropagation[3,7,30].

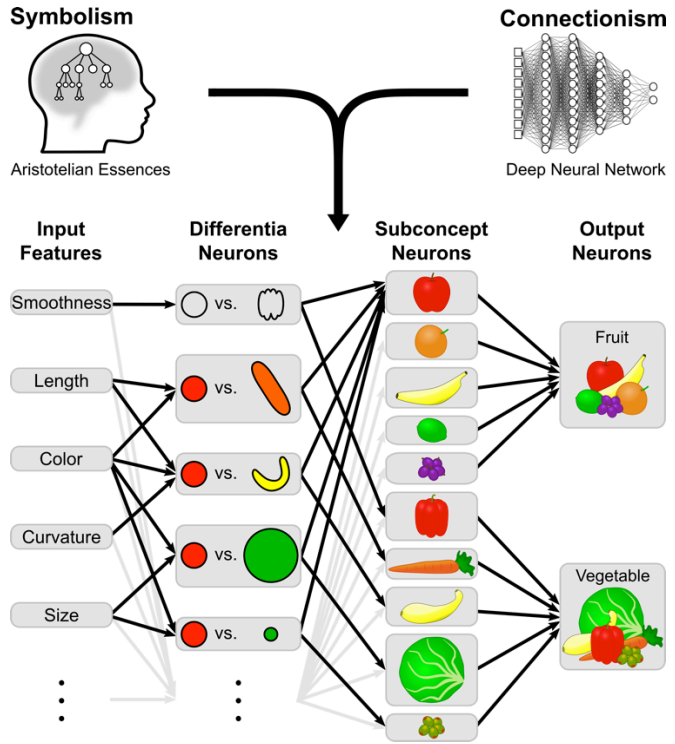

**Fig. 1. ENNs integrate symbolism and connectionism.** The essence model integrates the symbolic Aristotelian philosophy of essences with deep neural networks to produce essence neural networks (ENNs). This schematic shows how an ENN might classify a piece of produce as either a fruit or vegetable. The input features feed into differentia neurons, each of which distinguishes a particular fruit from a particular vegetable (only apple-distinguishing differentiae are drawn). Each differentia neuron innervates the appropriate subconcept neurons, which then feed into the corresponding output concepts.

### ENN performance is comparable to that of gradient descent-trained networks

ENNs are most similar in design to Voronoi neural networks (VNNs), which perform Voronoi tessellation on all training points individually and then combine them with AND and OR gates[31]. The key distinction is that ENNs learn from aggregate concepts and subconcepts, a fundamentally different approach that focuses on concepts instead of memorizing every past experience (i.e. exemplars) and therefore makes smooth, natural distinctions. One effect of this difference is that ENNs, in contrast to VNNs, can scale to problems with larger datasets and with many more features without having to grow exponentially in the number of neurons or in training time.
3

| Problem | Training Samples | Output Classes | Input size | Conv. Layers | Layer 1 neurons | Layer 2 neurons | GDN time (min) | ENN time (min) | GDN error | ENN error |
|---|---|---|---|---|---|---|---|---|---|---|
| **Perception tasks** | | | | | | | | | | |
| Rectangles | 50000 | 2 | 28x28 | — | 201 | 56 | 4.7 | 13.4 | 0.02% | 0.22% |
| Convex shapes | 50000 | 2 | 28x28 | — | 311 | 80 | 17.4 | 46.3 | 7.06% | 6.93% |
| MNIST | 60000 | 10 | 28x28 | — | 394 | 60 | 39.8 | 22.5 | 1.62% | 2.73% |
| MNIST | 60000 | 10 | 28x28 | 6 (5x5) 16 (5x5) | 127 | 84 | 41.5 | 61.0 | 1.03% | 2.35% |
| MNIST | 60000 | 10 | 28x28 | 32 (3x3) 64 (3x3) | 3167 | 84 | 49.8 | 123.5 | 0.93% | 0.86% |
| **Symbolic manipulation tasks** | | | | | | | | | | |
| Logic | 64 | 2 | 18 | — | 4 | 4 | 0.06 | 0.01 | 0% | 0% |
| Orientation: lines | 56 | 2 | 28x28 | — | 784 | 56 | 0.20 | 0.05 | 7.30% | 0% |
| Orientation: diagonal lines | 56 | 2 | 28x28 | — | 784 | 56 | 0.20 | 0.05 | 31.7% | 0% |
| Orientation: box outlines | 56 | 2 | 28x28 | — | 784 | 56 | 0.20 | 0.05 | 32.8% | 0% |
| Traveling salesman problem | 90 | 10 | 55 | — | 405 | 90 | 1.15 | 0.07 | 2.04 units | 0.00 units |
| Binary decision trees | 20 | 10 | 1024 | — | 180 | 20 | 0.05 | 0.02 | 0.794 nodes | −0.001 nodes |

**Table 1.** The results of training ENNs and a GDN of the same size on several datasets (Appendix A). Shown are the sizes of each training set, convolution filter number and size in convolutional layers, layer sizes for the networks, training times, and performance results on the test set for the ENN and its comparison GDN of the same size.

We used several datasets to train ENNs, each with a comparison gradient descent-trained network (GDN) of the same size (Appendix A), and found that ENNs and GDNs had comparable accuracy (Table 1 and Fig. 2). Without code optimization or parallelization, we found that ENNs require comparable training time as GDNs (Table 1 and Fig. 2), depending on ENN and GDN hyperparameter choices. In addition, the number of training samples actually contributing to ENN learning as support vectors grows sub-linearly with the total number of training samples (Fig. 2).

In order to demonstrate the potential to extend ENNs to more advanced network architectures, we have proposed one way to train convolutional ENNs (cENNs) (Appendix A). Sub-image windows are clustered to find local feature "subconcepts", and then SVMs are computed to generate the cENN's convolutional filters (Fig. 3). At the end of the convolutional layers a regular ENN is trained to produce the final output (Table 1).

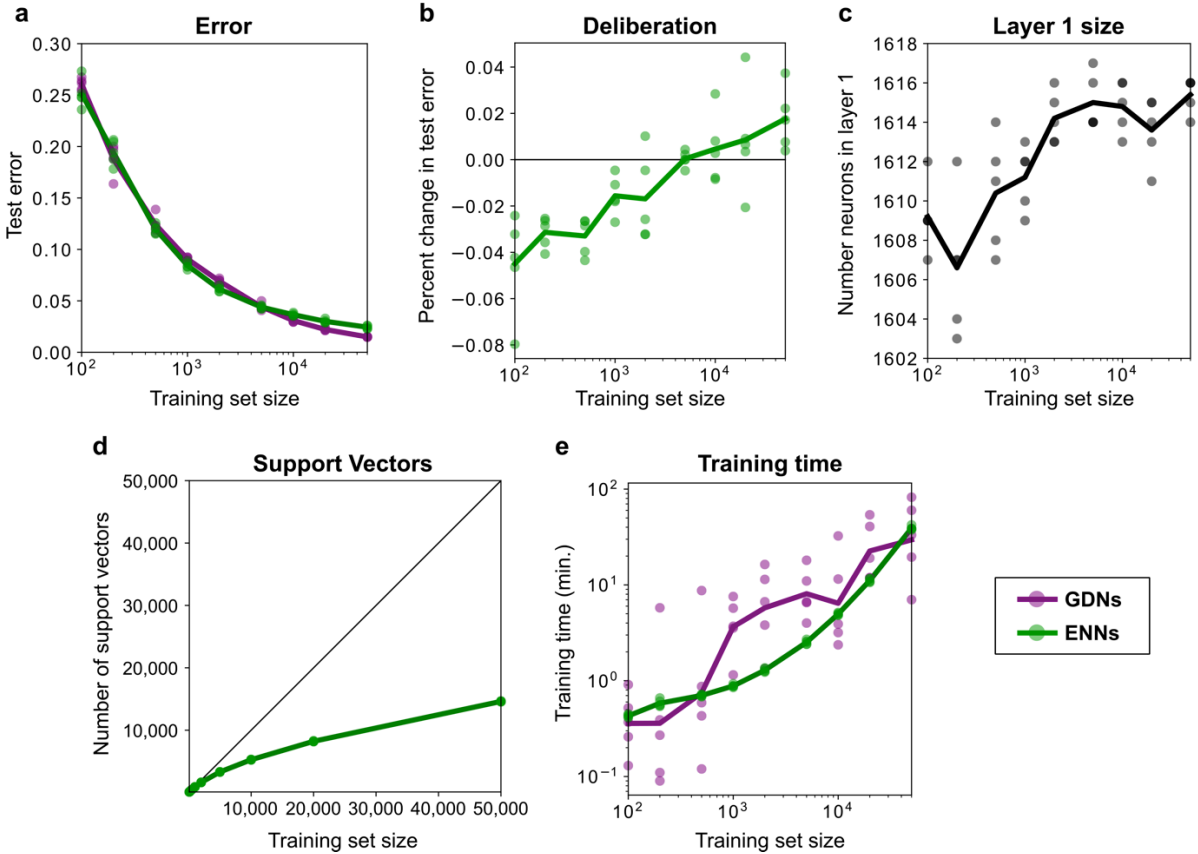

**Fig. 2. ENNs scale and perform well.** ENNs and GDNs trained on subsamples of the MNIST training set, tested on the full test set. Points indicate individual training runs, and the lines connect the average at each training set size (or the geometric average in (e)). **(a)** The test error for GDNs and ENNs with different training set sizes. **(b)** ENNs trained on smaller training sets typically improved with post-training deliberation (dENNs). **(c)** The networks each had 60 subconcept neurons, and without pruning differentia neurons from the first layer the number of neurons was approximately constant. **(d)** The number of training images that serve as support vectors grows sub-linearly with the training se size. **(e)** The training time for ENNs and GDNs was comparable across various training set sizes.

## ENNs are explainable

ENNs are transparent and explainable by construction because every neuron is designed for the specific purpose of separating opposing concepts. To demonstrate this, we trained networks with the popular MNIST dataset[8] of 70,000 images of handwritten digits (Fig. 4a) and with a synthetic dataset of 60,000 images of horizontally or vertically oriented rectangles (Fig. 5a). GDNs show little to no intelligible structure in the weights of synapses between image pixels and first-layer neurons (Fig. 4c and Fig. 5c), while ENN differentia neurons positively weight pixels more associated with a particular subconcept and negatively weight those of a different subconcept (Fig. 4b-c and Fig. 5b-c).

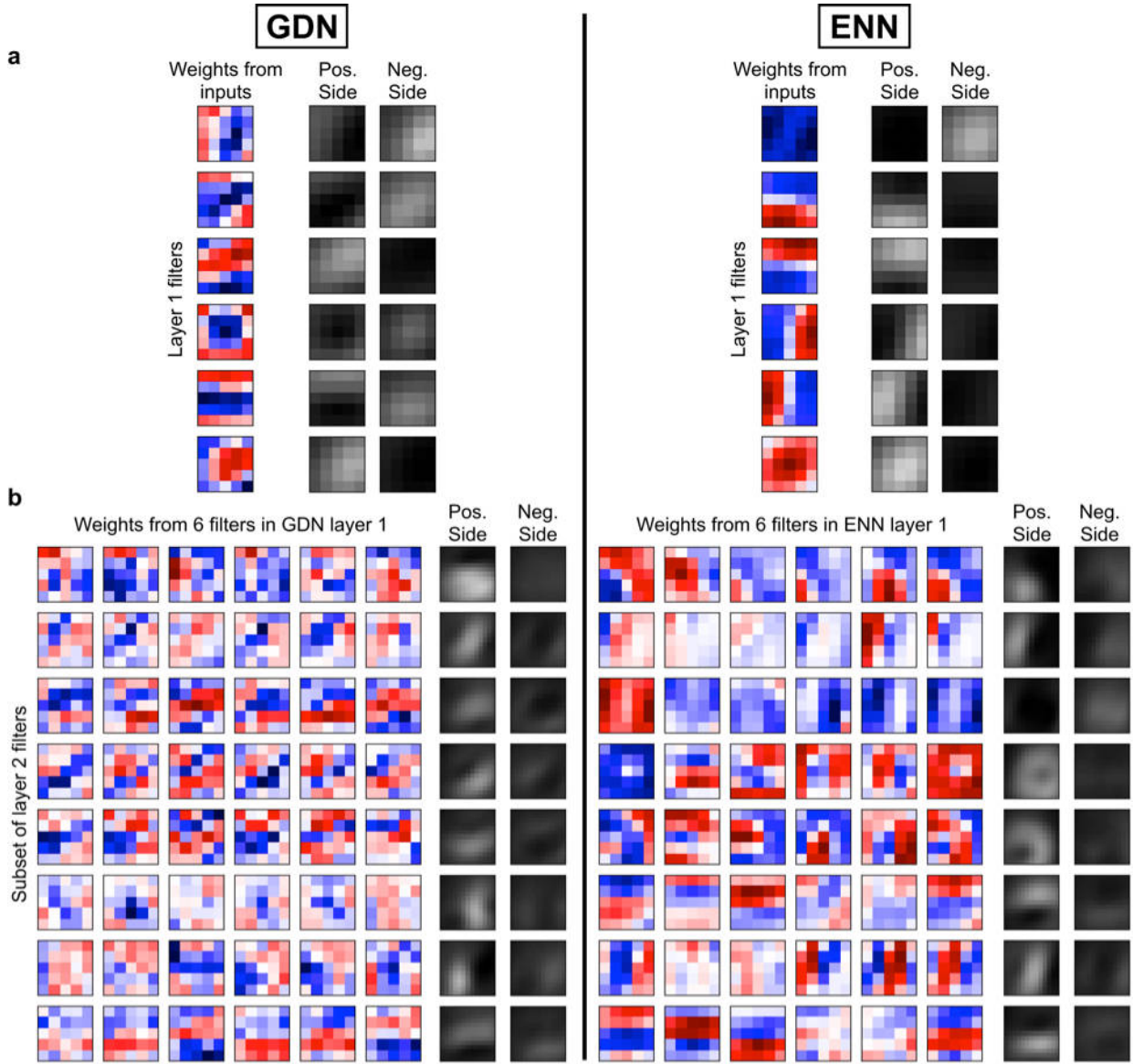

**Fig. 3. Convolutional ENNs learn more explainable features.** A convolutional ENN and GDN were trained on the MNIST dataset. They contained 6 5x5 filters in the first layer and 16 5x5 filters in the second layer. Each filter's weights are shown (red pixels are positive, blue are negative). Also shown are the weighted averages of image windows that fall on the positive and negative sides of each filter's hyperplane (see Appendix A). **(a)** The convolutional filters from the first layer. **(b)** Eight of the filters from the second layer. In both layers, the ENN seems to have more regular and interpretable filters, and the corresponding feature visualizations appear more discernible, especially those lying on the positive side of the filter.

Connections between deeper layers of neurons in GDNs are typically even less decipherable, with no natural sparsity[14] or modularity[15] (Fig. 4d-e and Fig. 5d-e). When the ENNs learned the weights between differentia and subconcept neurons, each subconcept was allowed to make use of all available differentiae. We observed though that they each chose to rely heavily on only a handful of differentiae, with relative sparsity of strongly weighted connections (Fig. 4d-e and Fig. 5d-e), corresponding to weight distributions with fat tails (Fig. 4f and Fig. 5f).

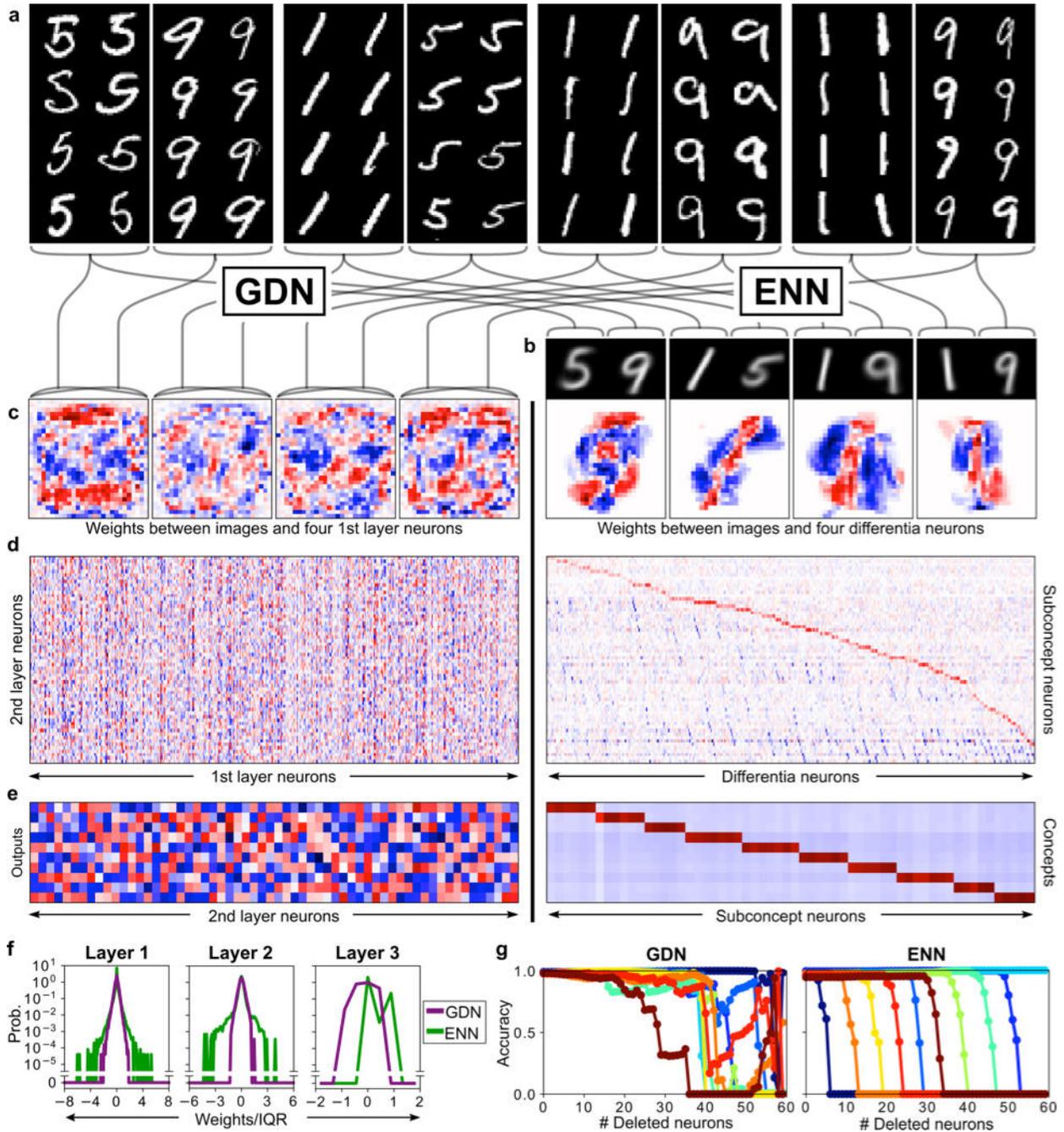

**Fig. 4. ENNs are explainable, sparse, and modular. (a)** Example MNIST training images. **(b)** ENN learning first clusters images within each concept into subconcepts. Several representative subconcepts are shown, with lines connecting from individual images to their subconcept average. **(c)** First-layer neurons, with incoming synaptic weights from each pixel of the input (red are positive weights, blue are negative). The ENN neurons shown are those that distinguish the pairs of subconcepts in (b). The GDN neurons shown are those that happen to maximally separate the ENN's subconcepts. **(d)** The connectivity matrix between the first and second layers of neurons. **(e)** The connectivity matrix between the second and third layers of neurons. **(f)** The distributions of synaptic weights from (c-e), showing fat tails for ENN weights. **(g)** The networks' accuracy in classifying each of the 10 classes of digits, colored separately, as subconcept neurons were sequentially deleted.

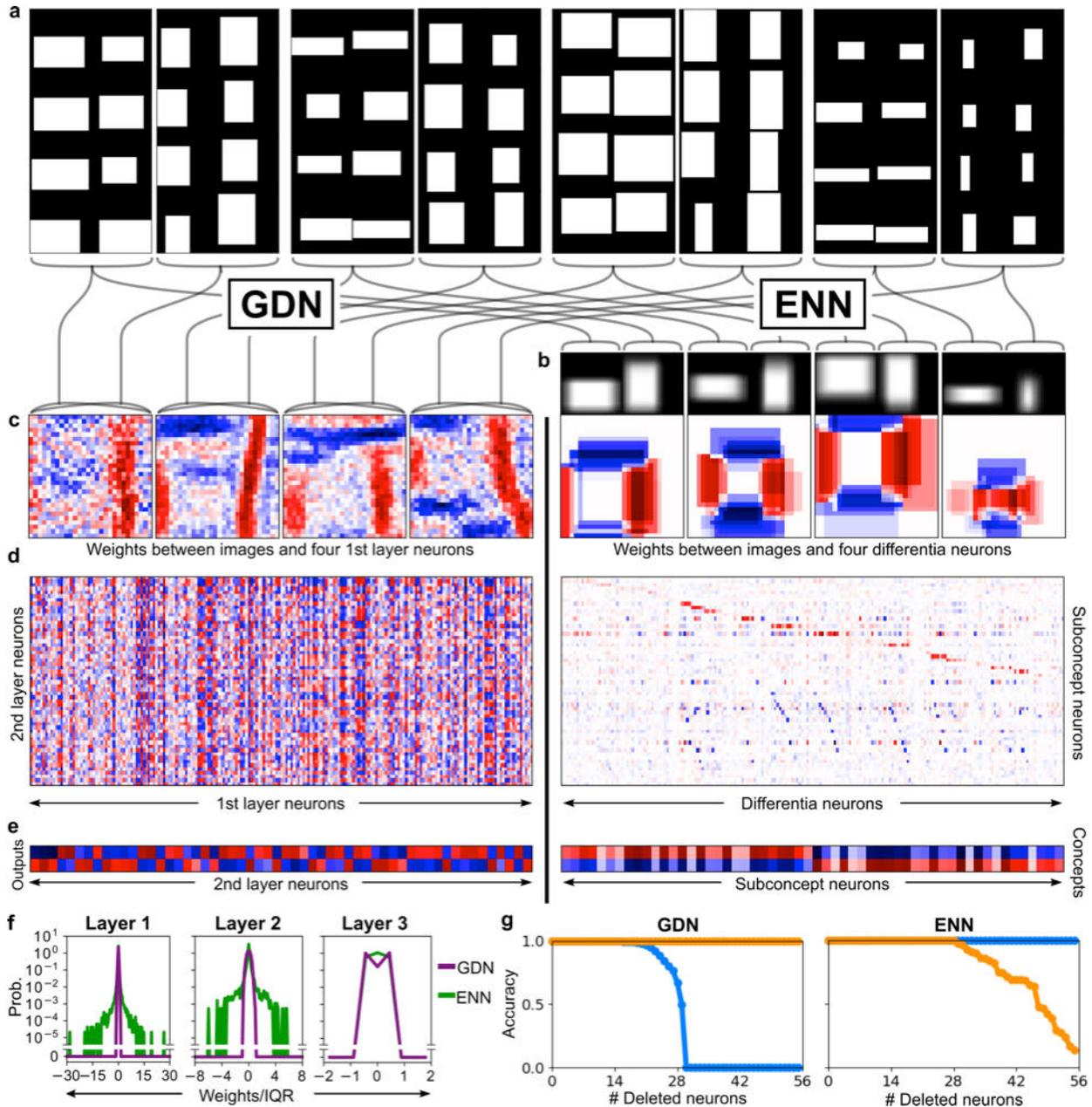

**Fig. 5. ENN weights are explainable, sparse, and modular. (a)** Example rectangle training images. **(b)** ENN learning first clusters images within each concept into subconcepts. Several representative subconcepts are shown, with lines connecting from individual images to their subconcept average. **(c)** First-layer neurons, with incoming synaptic weights from each pixel of the input (red are positive weights, blue are negative). The ENN neurons shown are those that distinguish the pairs of subconcepts in (b). The GDN neurons shown are those that happen to maximally separate the ENN's subconcepts. **(d)** The connectivity matrix between the first and second layers of neurons. **(e)** The connectivity matrix between the second and third layers of neurons. **(f)** The distribution of synaptic weights from (c-e), showing a fat tail for ENNs. **(g)** The networks' accuracy in classifying the two rectangle classes, colored separately, as subconcept neurons were sequentially deleted.

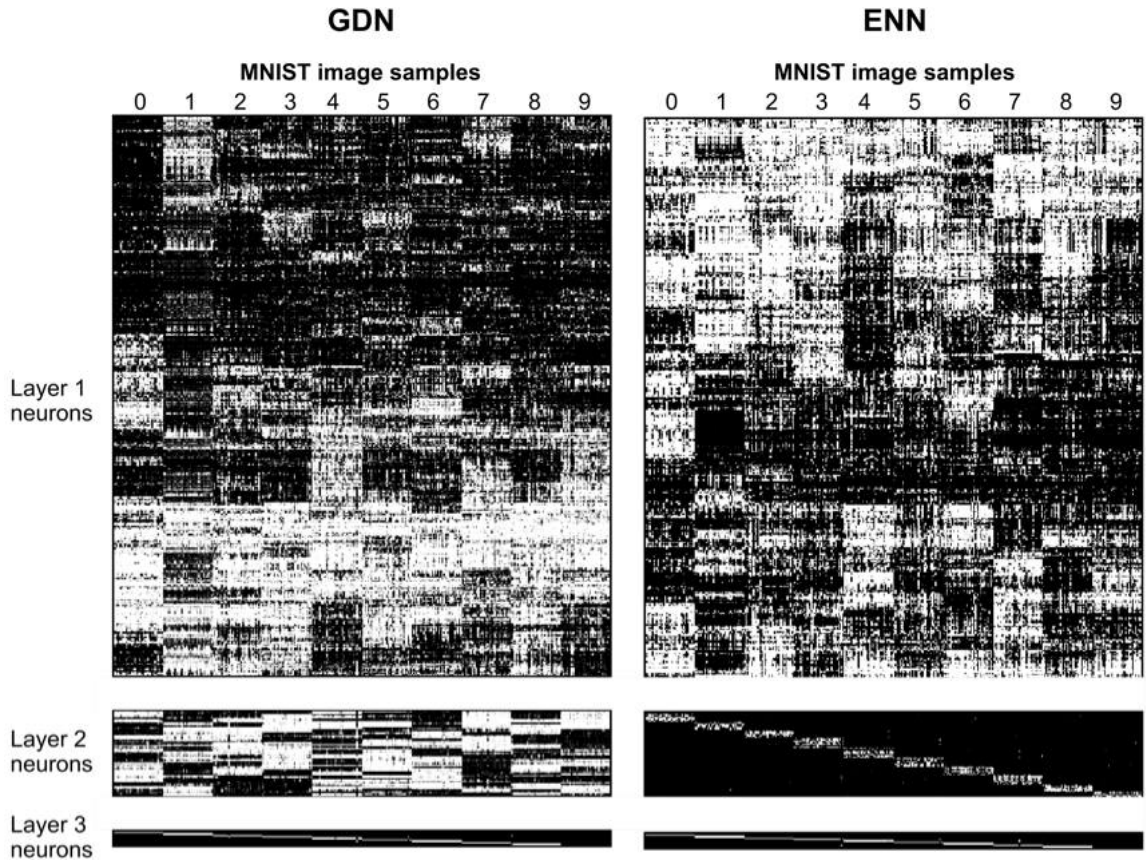

**Fig. 6. ENN neurons use distributed and localized firing patterns.** The firing patterns of all neurons in a GDN and ENN, trained on the MNIST dataset, when ex-posed to 350 random test images of different digits. Neuron outputs are shown in grayscale, with maximal firing in white and zero firing in black. The GDN and ENN both appear to have distributed firing patterns, but ENNs are able to utilize more sparse, localized firing in the sec-ond layer. Neurons were arranged by cluster analysis on the firing patterns across these 350 test images. The networks used sigmoid output neurons to better mimic biological neural networks.

The sparsity in the last two ENN layers suggests a higher degree of modularity, which, though not explicitly enforced during training, is the expected consequence of the ENN model (Fig. 1). This functional modularity is apparent when we progressively delete second-layer (sub-conceptual) neurons and observe loss of class-specific accuracy that is sequential in ENNs and haphazard in GDNs (Fig. 4g and Fig. 5g). Modularity is a key feature of biological neural networks, which is why sequential loss of function is also observed in progressive neurological disorders with focal lesions, such as multiple sclerosis[32] and vascular dementia[33]. It is the first example, to our knowledge, of an artificial neural network in which sparsity and modularity arise naturally without being purposely designed.

The explainable nature of ENNs also makes them amenable to post-training adaptation. We focused on simulating the two cognitive systems described by dual-process theory, System 1 making rapid, intuitive decisions and System 2 performing slow, deliberative reasoning[34]. Feed-forward neural networks are already analogous to System 1, and we found one way to mimic System 2 is to allow ENNs to become deliberative (dENNs), dynamically modifying the strictness of subconcept neurons whenever there was low classification certainty (Appendix A). This often offered improvement in classification accuracy, especially when training with less data (Fig. 2) or on symbolic problems (see below). The black-box nature of GDNs, however, precludes simulation of System 2.

Furthermore, the explainability of ENN neurons leads to functional interpretability of ENN firing patterns. Mental representations in the brain have been described as either distributed or localized. While a distributed representation is spread out in firing patterns across many non-selective neurons, a localized representation involves neurons highly selective for specific stimuli or processes (e.g. "grandmother cells")[35,36]. We observed that ENN differentia neurons demonstrate a distributed firing pattern, while subconcept and concept neuron firing is sparse, localized, and selective (Fig. 6). Such hierarchical separation of distributed versus localized firing patterns is also observed in numerous animal nervous systems[37,38].

### ENNs can generalize concepts using symbolic manipulation

ENNs are explainable and can support localized firing patterns because each neuron encodes a fuzzy symbol, i.e. similarity or dissimilarity to opposing concepts. Since symbolic thought is the basis of human reasoning, we asked if ENNs are capable of more explicit symbolic manipulation at the neuronal level. We defined this as having neuron outputs of only 0 or 1 (or 0.5 in the case of uncertainty or ties), which is consistent with spike-based theories of neuron signal coding[39] and the practice of one-hot encoding categorical variables. GDNs cannot achieve this due to the need for differentiable neuron outputs.

We trained a GDN and an explicitly symbolic ENN to simultaneously learn all 16 two-input Boolean functions (e.g. AND, OR, NAND, XOR) by training on all 64 entries of the truth table (Fig. 7a). Each training sample was a vector with 18 values, the first two representing the function's binary inputs and the remaining 16 serving as a one-hot encoding of the Boolean function itself. Both GDNs and ENNs were successful at achieving perfect accuracy on this dataset (Table 1). We made ENNs learn large-magnitude weights so that each neuron produces a symbolic output of 0 or 1 (Fig. 7b), and we again found that the ENN weights, but not the GDN's, were regular and explainable (Fig. 7c). This allowed us to map the ENN directly to a logic circuit (Fig. 7d), demonstrating the capacity for ENNs to implement symbolic manipulation at the network level using symbolic neurons.

One purpose of symbolic reasoning is to learn simple rules from limited experience and apply them to more complex problems. We define this blind generalization as the ability to extrapolate from one distribution of inputs to a different one without any additional training. To test this, we had a GDN and a symbolic ENN learn horizontal versus vertical orientations on a set of images each with a pixel-wide white stripe, totaling 56 28x28 images (Fig. 7e). The ENN had symbolic firing (Fig. 7f) and was therefore able to be translated into pseudocode (Appendix B). Moreover, unlike the GDN, the ENN could correctly describe the orientation of shorter line segments, diagonal line segments, and rectangular box outlines with zero blind generalization error (Table 1 and Fig. 7g). We tried optimizing GDNs by changing their size and choosing networks with the best test set performance, but still could not find one that generalized perfectly (Fig. 8 and dotted lines in Fig. 7g). To see how difficult it is for gradient descent to train a blindly generalizing network, we trained GDNs seeded with ENN weights plus added noise. Gradient descent could no longer train a network that perfectly generalized to the diagonal line and box outline images once the weight perturbations exceeded 1% and 3%, respectively (Fig. 9).

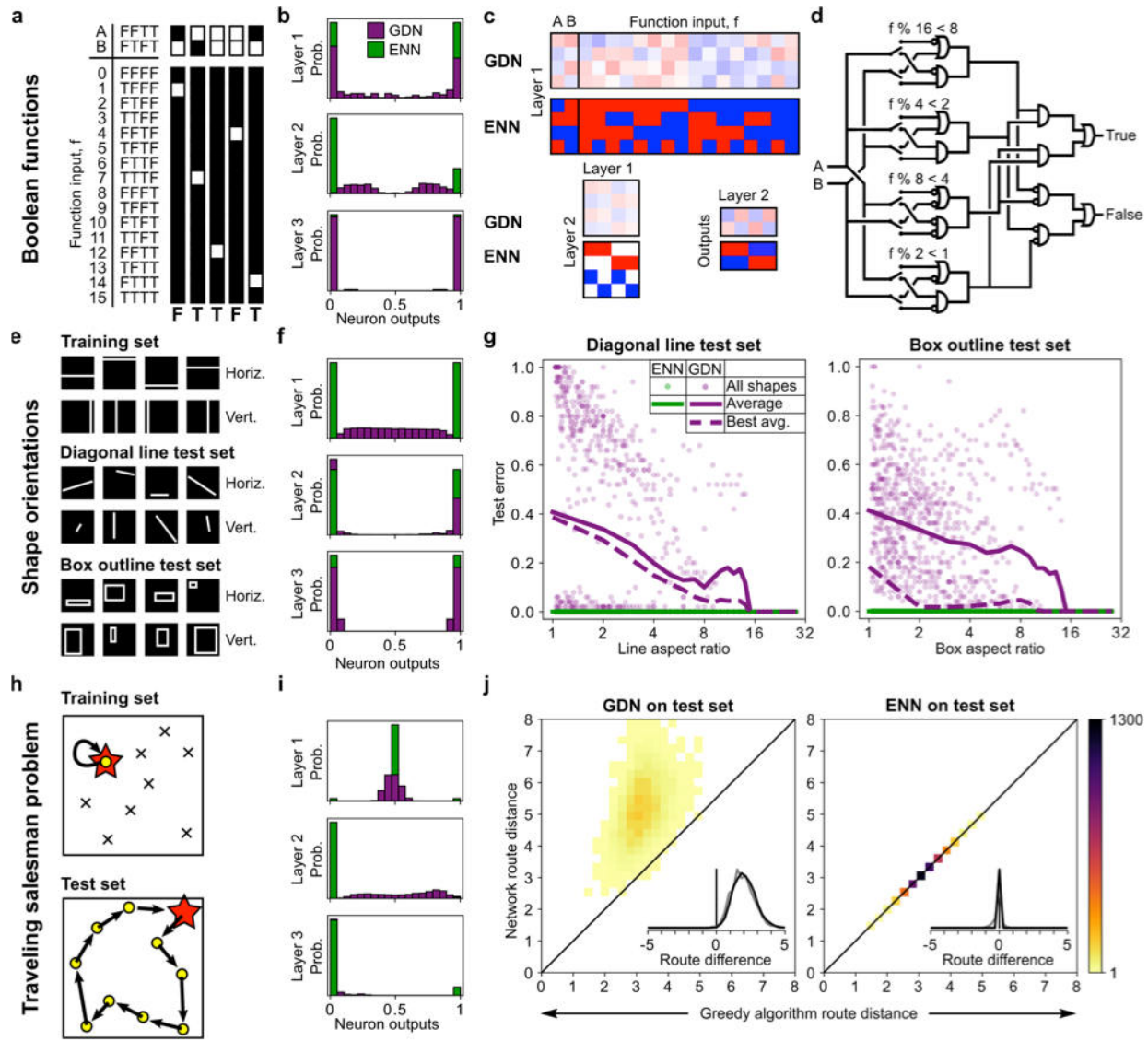

**Fig. 7. ENNs can implement symbolic manipulation and generalize rules. (a-d)** Boolean logic results. **(a)** The truth table for all 16 functions, with 5 examples of entry encodings. **(b)** Distributions of neuron outputs across all inputs. **(c)** The connectivity matrices of GDN and ENN layers (red are positive weights, blue are negative). **(d)** The ENN-derived logic circuit. **(e-g)** Orientation problem results. **(e)** The training set contains images with an image-wide stripe, while the test sets include either diagonal lines or box outlines. **(f)** Distributions of neuron outputs on the orientation training set. **(g)** While the ENN generalized perfectly to the diagonal and box datasets, the GDN cannot, especially for lower aspect ratios. Each point represents the classification error of a particular diagonal or box in different locations across the image. **(h-j)** TSP results. **(h)** Our maps consisted of 10 cities. The TSP is to find the shortest route from the red star to all cities (yellow dots). Black x's are previously visited cities. **(i)** Distributions of neuron outputs on the TSP training set. **(j)** A comparison of the distances of routes found by the nearest-neighbor algorithm to the GDN and ENN. Inset is the distribution of differences in routes (black lines), including results from expanding the training set (gray lines).

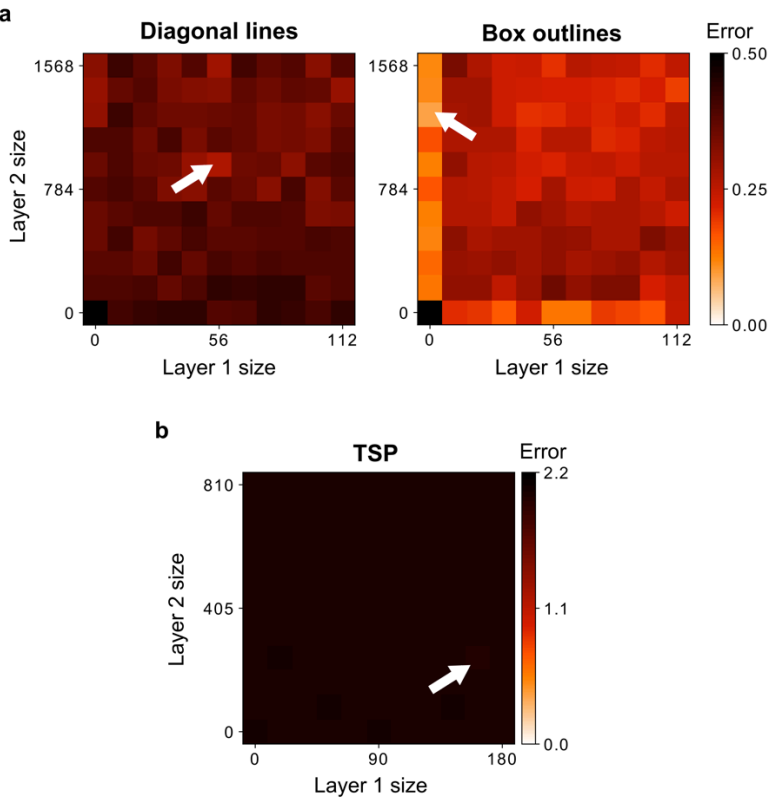

**Fig. 8. GDNs of different network sizes cannot generalize.** We trained GDNs on the training sets of the orientation problem and the TSP and tried to optimize hyperparameters that gave the best performance on the generalized test sets. Shown here are the results of a grid search over network sizes. Arrows indicate the network size that produce the best possible GDNs. **(a)** Heatmaps of the average test error of 5 orientation-trained GDNs of different sizes on the diagonal line and box outline test set sets. **(b)** A heatmap of the differences in route distances found by TSP-trained GDNs and the greedy algorithm for different GDN sizes, averaged over 5 trained GDNs.

Next, we asked if neural networks trained on a simple, non-image problem could learn a greedy rule transferable to more complex problems they had never seen. We did this first for the traveling salesman problem (TSP), a classic NP-hard problem[40]. The training set had only 90 10-city maps, each starting from a different city and with only one unvisited city remaining, located at zero distance from the current one (Fig. 7h). The trained ENN was again symbolic (Fig. 7i), required deliberation (dENN) to break ties, and was also translated into pseudocode (Appendix B). The GDN, despite perfect training accuracy, did not generalize to the full-map test set, while the ENN performed identically to a greedy algorithm using the nearest-neighbor heuristic (Fig. 7j). Searching for a different network size again showed no improvement (Fig. 8). Moving 4000 of the 5000 test problems into the training set yielded only a slight improvement for both GDNs and ENNs (gray lines in Fig. 7j). We tried again to seed GDNs with ENN weights plus noise, but the GDNs could not reproduce the dENN success with any amount of noise or number of training epochs (Fig. 9).

We replicated these results on the optimal binary decision tree (BDT) problem, which is also NP-hard[41]. At each branch point of the tree the networks were asked to choose which feature to split. Training included only the 20 10-feature truth tables for which the optimal BDT contained a single branch node, while the test set included truth tables with deeper optimal BDTs (Fig. 10a). We found similar results as the TSP (Fig. 10b-c and Appendix B), with dENN performance on par with the greedy CART algorithm.

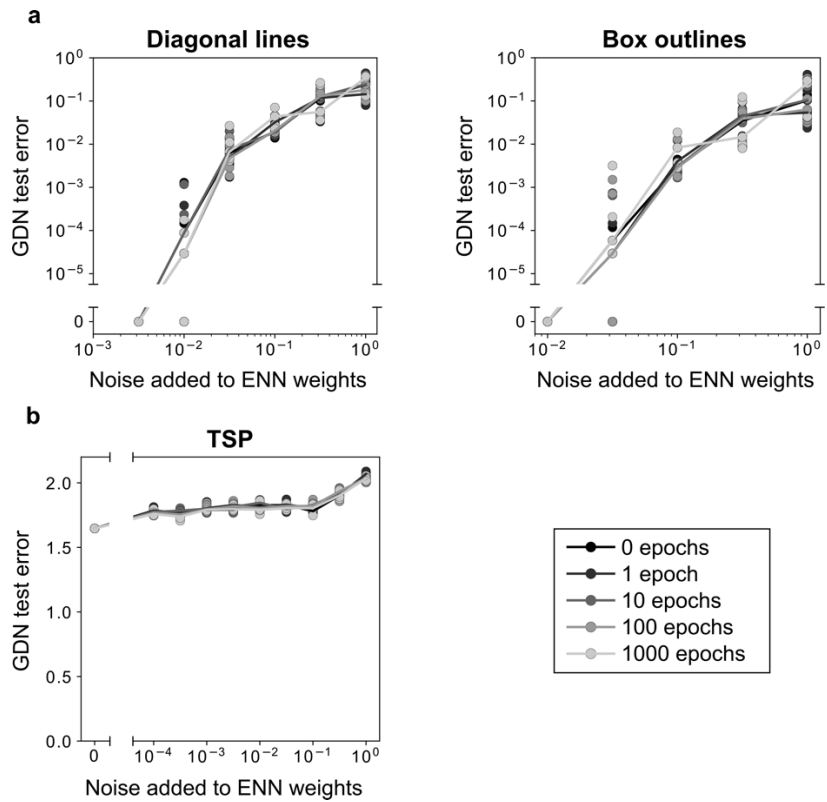

**Fig. 9. Improbability of gradient descent training blindly generalizing networks.** GDN pre-training weights were seeded with ENN weights that had various amounts of noise added to them. The results of these GDNs trained for different numbers of epochs are shown. Dots indicate all 5 repeats for each noise level and number of epochs. **(a)** The test error of orientation-trained GDNs on the diagonal line and box outline test sets. **(b)** The average distance in route lengths found by the TSP-trained GDN and the greedy nearest-neighbor algorithm.

**Fig. 10. ENNs can learn a greedy symbolic algorithm to build BDTs.** **(a)** Simplified schematic of the BDT problem for 3-input truth tables (the BDT dataset used 10-input tables). Networks were trained on truth tables that only require a one-node BDT and were tested on truth tables requiring much deeper BDTs. **(b)** The ENN was able to learn neurons that fire symbolically while the GDN did not. **(c)** A heatmap for all 5000 test-set truth tables comparing the depth of BDTs generated by the greedy CART algorithm or either the GDN or ENN. The inset shows the distribution of differences in the tree depths found, with the gray lines behind showing the results when test set samples were transferred to the training set (not visible for the ENN).

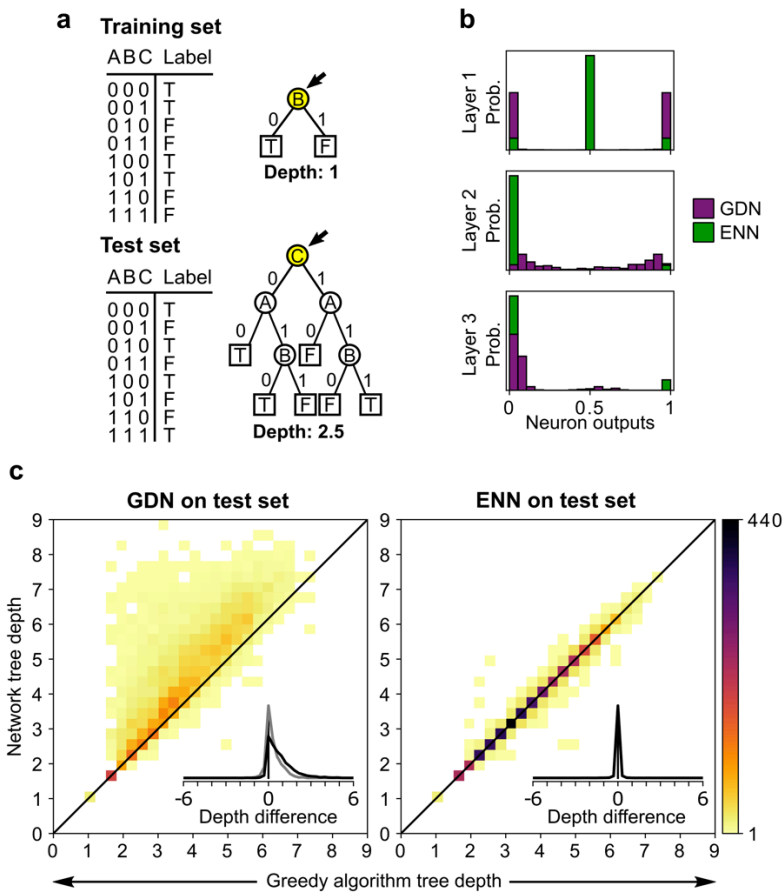

To our knowledge, these are the first instances of blind generalization by artificial neural networks. Previous work with domain adaptation has assumed the inclusion of at least some elements from the new distribution[42], while ENNs can learn generalizable rules from the simpler distribution alone.

**ENNs are more robust to input noise and adversarial attacks**

In order to better understand the differences in decision-making between GDNs and ENNs, we inspected the decision boundaries learned by the networks. We can directly visualize them from the Boolean logic problem for various Boolean functions (Fig. 11a). The ENN had more intuitive decision boundaries, which is the result of using SVMs in ENN learning to space decision boundaries evenly. Despite training on only binary function inputs, the ENN interpolated well on fuzzy inputs, in contrast to the GDN's inconsistent and arbitrary interpolation (Fig. 11b).

To measure the separation between inputs and decision boundaries on less structured problems, we took individual images from the MNIST and rectangle test sets and interpolated between them and either an image of a different class or white noise. Along this interpolation we found the location of the nearest decision boundary and measured the normalized $L_1$ distance (i.e. the average pixel difference) from the starting image. We found that both GDNs and ENNs space decision boundaries between images at about the same distance (Fig. 11c and Fig. 12a). Howev-er, when interpolating between images and white noise we observed ENN decision boundaries spaced at a greater distance than those of GDNs, suggesting a more robust placement of decision boundaries.

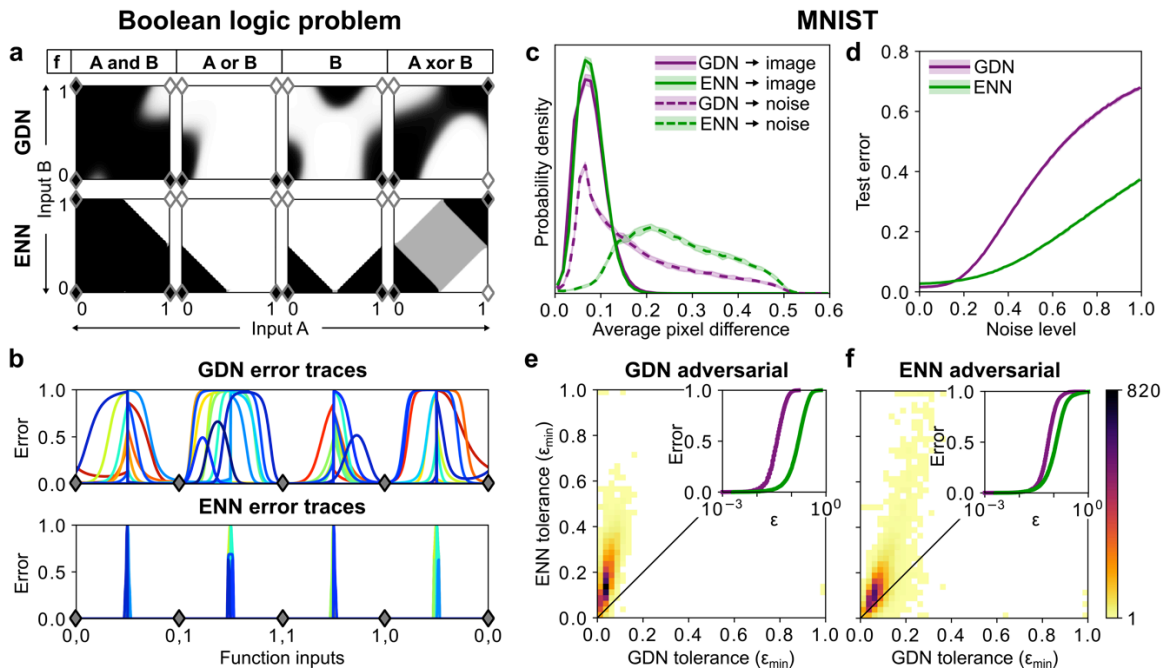

**Fig. 11. ENN decision boundaries confer robustness to noise and attacks. (a)** A GDN and ENN trained simultaneously on all 16 Boolean logic functions, with 4 shown here. A and B are function inputs, and in grayscale is the networks' output probability for True. Diamonds indicate training samples, with True in white and False in black. **(b)** The error between the network output and the closest corner while tracing around the unit squares from (a), each of the 16 Boolean functions shown in different colors. **(c-f)** A GDN and ENN trained on MNIST, with results on the test set. **(c)** The probability distributions of the average pixel difference between images and a decision boundary, found by interpolating toward either another test images or white noise (with interdecile range shaded). **(d)** The network classification error as Gaussian noise is added to images (with interdecile range shaded). **(e-f)** Adversarial attacks were generated against both the GDN and ENN for all test images, with the minimum tolerated $\varepsilon_{min}$ scaling factors shown in these two heatmaps. Inset is the misclassification error at different $\varepsilon_{min}$ values.

This meant that the ENNs had a greater tolerance to input noise than did GDNs (Fig. 11d and Fig. 12b). Moreover, robust decision boundary arrangement is particularly important when defending against adversarial attacks. We generated adversarial images against GDNs and ENNs using the fast gradient sign method[11] and measured the minimum perturbation ($\varepsilon_{min}$) needed for each image to fool its network. Because adversarial images transfer well between networks[11], we also tested each network on adversarial images designed against the other. Not only were ENNs several-fold more robust to self-adversarial images than were GDNs, but they were also not fooled by transferred adversarial attacks designed against GDNs (Fig. 11e-f and Fig. 12c-d). On MNIST we also found that the GDN was less robust to attacks designed against the ENN than was the ENN itself. Furthermore, this difference in robustness to adversarial attacks was even greater when training larger networks (Fig. 12e-h). Interestingly, we also observed that the ENN adversarial perturbations appeared more interpretable (Fig. 13). These results indicate that ENNs are naturally more robust to adversarial attacks.

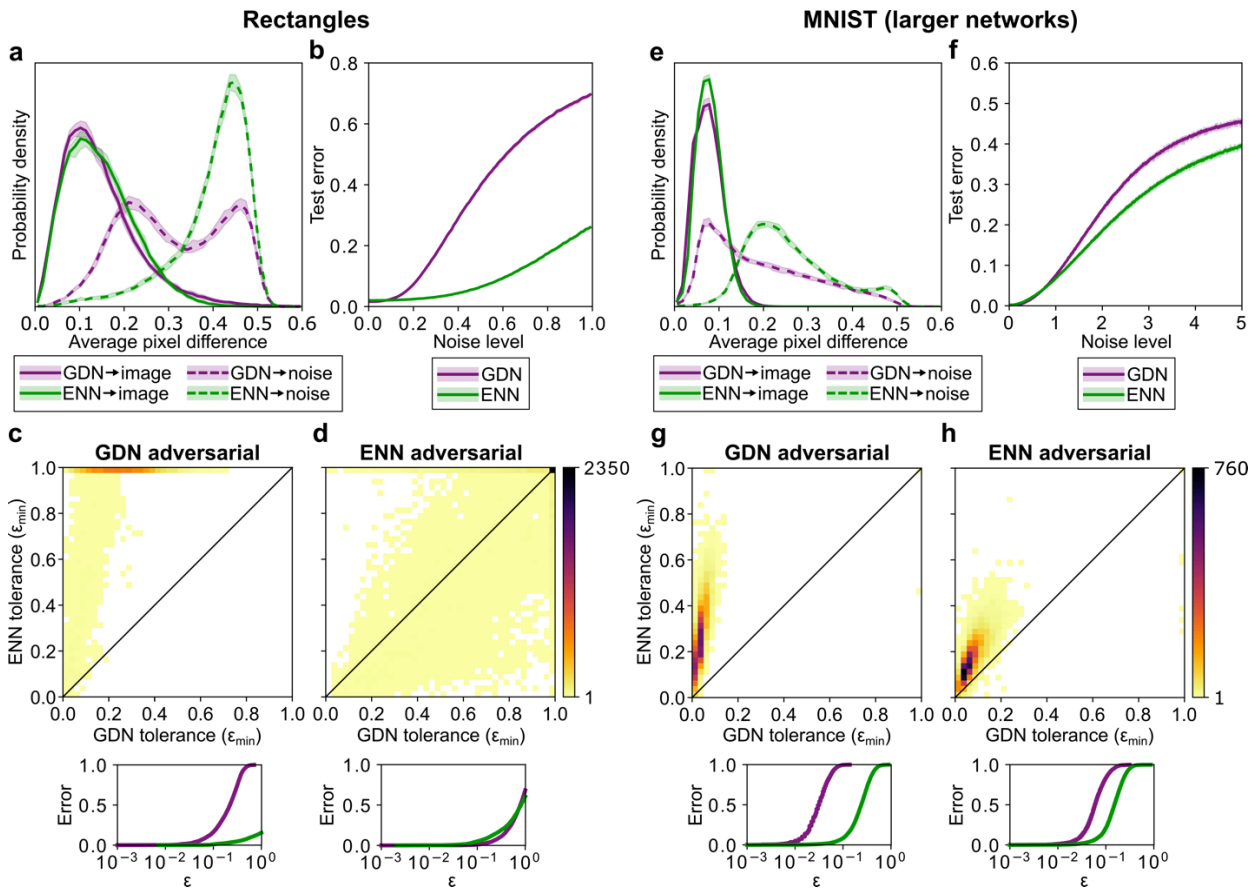

**Fig. 12. ENNs are robust to noise and adversarial attacks.** (a-d) Results for a GDN and ENN trained on the rectangle problem. (e-h) Results for a GDN and ENN trained on MNIST that are larger than in Fig. 11 (1465 differentiae and 150 subconcepts). **(a, e)** The probability distributions of the average pixel difference between images and a decision boundary, found by interpolating toward either another test images or white noise (with interdecile range shaded). **(b, f)** The networks' test error with increasing amounts of Gaussian noise added to images (interdecile range shaded). **(c-d, g-h)** Adversarial attacks were generated against both the GDNs and ENNs for all test images, with the minimum tolerated $\varepsilon_{min}$ scaling factors shown in these two heatmaps. Below are the error rates at different $\varepsilon$ values.

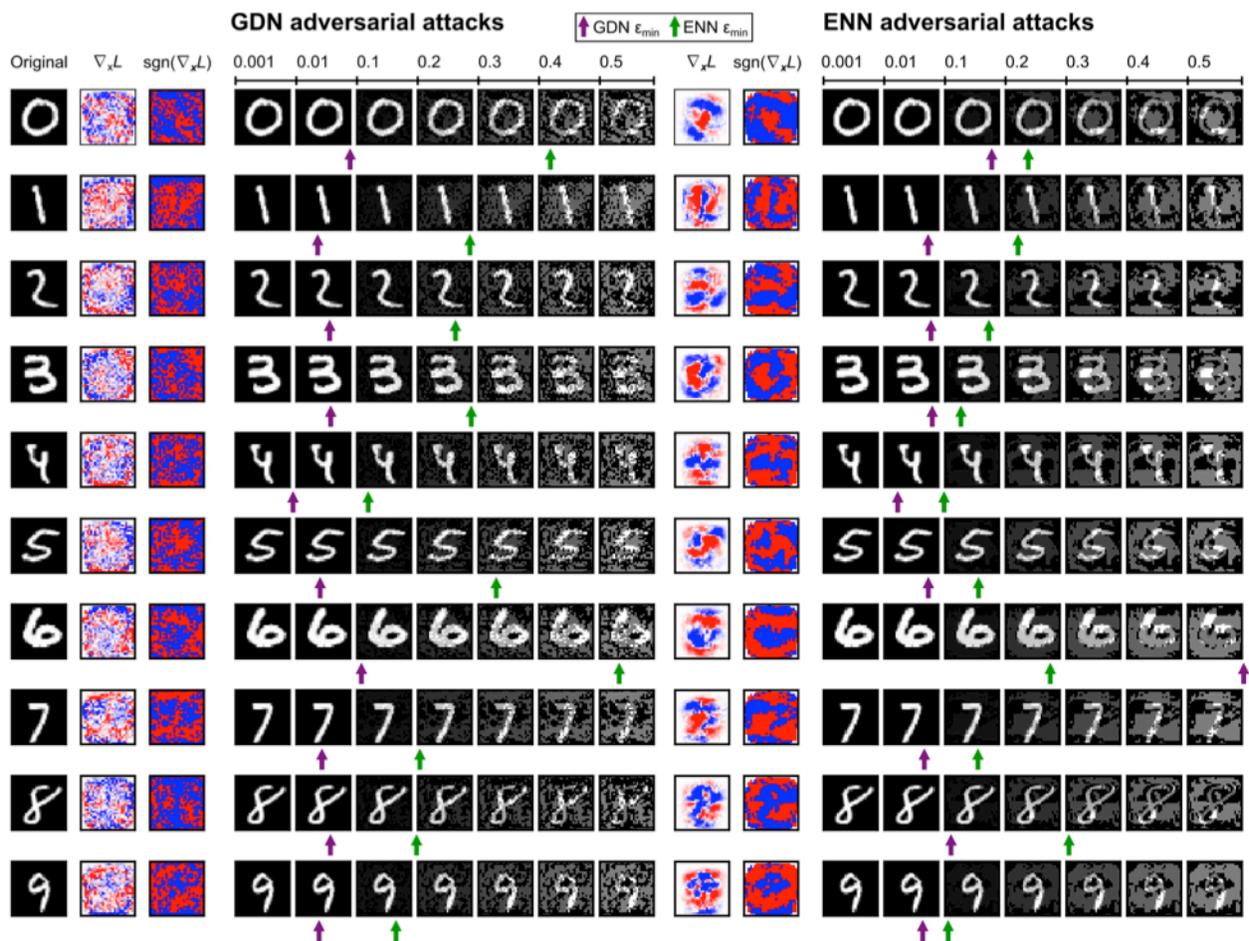

**Fig. 13. Examples of adversarial attacks.** Adversarial attacks were generated against both GDNs and ENNs. Shown are examples from the MNIST-trained networks from Fig. 12g-h. The gradient of the loss function $\nabla_x L$ is shown, as is $\text{sgn}(\nabla_x L)$, which is the adversarial perturbation being added (positive values are shown in red, negative in blue). This perturbation is scaled by an $\varepsilon$ value until the network no longer classifies the image correctly, which are shown for increasing values of $\varepsilon$. Underneath the image is an error showing the $\varepsilon_{min}$ at which either the GDN or ENN misclassified. Interestingly, the ENN adversarial perturbations are negative in regions specifically associated with the particular digit.

## Discussion

We have kept the ENN learning algorithm simple in order to demonstrate its power, which means there is great potential for improvement. Many possible improvements are technical or in line with current deep learning methods, including building wider and deeper networks, using unsupervised learning to find overlapping or hierarchical subconcepts, adding recurrent and skip connections, and implementing both online and transfer learning. But the symbolic and explainable nature of ENNs also makes possible fundamentally new types of improvements. We have introduced several already, showing that ENNs can deliberate, learn hierarchical structure from the data, use unsupervised learning with subconceptualization, and generalize concepts from a few instructive examples. Other new possibilities that ENNs should open include incorporating prior human knowledge, designing network architectures and hyperparameters rationally, learning by definition, and making causal inferences.

In several places we have shown how ENNs are consistent with neurobiological observations and theories, describing the topologies and firing patterns of well-studied biological neural networks. In Table 2 we have mapped several known neural circuits to the ENN model and have also included representative examples from biomolecular, cellular, and population systems that fit within the essence framework.

The key insight of the ENN model is that biological and artificial neurons make distinctions among possible inputs, which is then used by downstream neurons to make further distinctions. In an exclusively distributed network (modeled by GDNs), neurons make arbitrary, unstructured distinctions, so an entangled representation is propagated through the network until being disentangled at the end[11]. ENN neurons, however, make natural, common-sense distinctions between concepts to separate or unify them so that the propagated representation is not entangled. This is what allows ENNs to simulate reasoning (e.g. symbol manipulation) in addition to perception (e.g. image recognition). ENNs may thus provide a general framework for understanding how cognition can arise in biological neural networks and be simulated in artificial intelligence systems.

**Table 2. Biological systems interpreted in the essence framework.** For various types of biological systems, we have provided representative examples of how their constituents fit into the essence framework.

| System | Inputs | Differentiae | Subconcepts | Output(s) | Ref. |
|---|---|---|---|---|---|
| **Biomolecular** | | | | | |
| Kinetic proofreading | Substrate structure and chemistry | Transition and off rates | Exit points | Catalysis or signaling | 43 |
| Cytochrome p450 (e.g. CYP3A4) | Substrate structure and chemistry | Binding site structure and chemistry | Ligand-binding conformations | Oxidation | 44 |
| **Cellular/Organismal** | | | | | |
| Bone morphogenetic protein (BMP) pathway | BMPs | BMP receptors | SMADs | Growth and development | 45 |
| Immune response | Foreign particles or cells | Receptors of various immune cells | Activation of various immune cells | Clearance or neutralization | |
| **Neural circuits** | | | | | |
| *C. elegans* locomotion | 1-octanol and food | ASH and AWC neurons | AVA and AIB neurons | Backward motion | 46 |
| Insect olfaction | Antennal lobe glomeruli projection neurons | Kenyon cells of the mushroom body | Mushroom body output neurons | Downstream neuropils | 47 |
| Weekly electric fish electroreception | Electroreceptors | Pyramidal cells in electrosensory lateral line lobe | Torus semicircularis neurons | Electric perception | 48 |
| Rodent whisker sensation | Whisker mechanoreceptors | Brainstem barrelette neurons | Thalamic VPM barreloid neurons | Barrel cortex neurons | 49 |
| Rodent and human spatial perception | Sensory inputs | Grid and border cells | Place cells | Spatial perception | 50 |
| **Population-level** | | | | | |
| Ant and honeybee nestmate recognition | Incoming ant's cuticle | Individual ant interactions with another's cuticle | Nestmate intra-colony interactions | Aggression | 51 |

## Acknowledgments

**Funding:** The authors acknowledge the Cecil H. and Ida Green Foundation, the Welch Foundation and the Leland Fikes Foundation HR/HI grant for funding this research. **Author contributions:** M.M.L. oversaw the project. M.M.L. and P.J.B. initiated the research. P.J.B. developed the model, wrote the code, and performed experiments. P.J.B. and M.M.L. wrote the manuscript. **Competing interests:** None to report.

**Appendix A: Methods**

Code availability

Code used in this project will be made available at a future date. Networks were trained on the BioHPC computing cluster at the University of Texas Southwestern Medical Center.

ENN algorithm

There are different possible ways to train a neural network consistent with the ENN model, but we have chosen an effective yet simple approach that performs well. There are 5 modules in our ENN learning algorithm: (i) Learn subconcepts within each conceptual class, (ii) learn differentiae between each pair of subconcepts, (iii) prune unnecessary differentiae, (iv) unite all samples into each subconcept, and (v) unite each conceptual class.

(i) Finding subconcepts within each concept is done with unsupervised learning. We elected to use hierarchical linkage clustering within each class, choosing a single cutoff value for all concepts' linkage trees such that the desired total number of subconcepts was obtained. We found that the ward clustering metric gave the best results due to its emphasis on generating compact clusters of comparable size.

(ii) Learning differentiae between each pair of subconcepts was done using linear SVMs. In practice, we found it unnecessary to compute differentiae between subconcepts of the same concept. The weights and intercepts of the SVMs were all scaled by a multiplier hyperparameter to alter the steepness of the neuron response, and these became the weights and biases of the inputs to each differential neuron in the first layer. The activation function used was the standard sigmoid function, $\sigma(x) = \frac{1}{1+e^{-x}}$.

(iii-iv) In order to prune away differentiae, steps iii and iv were done together. For each subconcept, an initial SVM was generated between itself and all other concepts using the differentia neuron outputs. To improve running time, this SVM only used as features the differentiae associated with the particular subconcept. We sequentially masked neurons whose weight magnitudes in this SVM were low and then recomputed the SVM. This sequential pruning halted either when the SVM's margin dropped below a certain fraction of the original SVM's margin or when its misclassification error increased by a certain amount. Once this was done for each subconcept, those differential neurons that were no longer being used by any subconceptual SVM were pruned from the network. The final subconceptual neurons were generated with new SVMs that had access to all differential neurons. As before, the weights and intercepts of these SVMs were scaled by a multiplier and became the weights and biases for each subconceptual neuron, which was followed by a sigmoid activation function.

(v) To connect the subconceptual neurons with the output conceptual neurons, either an SVM was first computed for each conceptual class separating it from all other classes or each subconcept was simply connected to its own concepts. However, to assign more meaningful output probabilities, we refined these weights using a stochastic gradient descent approach for this final layer, using a categorical cross-entropy loss function. This gradient descent was also used to find the best value for the subconcept SVM multiplier hyperparameter. At the end of the ENN was placed a softmax layer to turn the concept neuron outputs into probabilities or optionally left as a sigmoid activation.

ENN hyperparameters

There are several hyperparameters found in this algorithm above that are user-defined. The number of subconcepts is the size of the second layer of neurons. Each SVM requires a cost to set the softness of the margin and a multiplier to scale the response. This multiplier was set to be very large for symbolic ENNs so that each neuron's output was either 0 or 1 (or 0.5 if the input was exactly zero). In the pruning step there is both a tolerated margin fraction and misclassification tolerance used to determine when to halt. With gradient descent in the final layer the multiplier for the subconceptual layer was also found, with the hyperparameter here serving as a maximum value.

In order to find an optimal set of hyperparameters, a grid search was done using 10-fold cross validation. To speed up the search process we did this on a restricted set of training data to narrow down several hyperparameters. For the final results, several ENNs were trained on each problem with all the same hyperparameters except for a variable number of subconcepts and a variable pruning toleration margin to vary the size of the network. We chose these parameters so that the network size would be generally comparable to previously published work[8,52].

Convolutional ENNs (cENNs)

We implemented a very simple yet effective method of training convolutional ENNs consistent with the essence model. Training convolutional ENNs begins by obtaining subimages from the training set randomly sampled equally from each class and uniformly within each image. *k*-means clustering was used to divide up the subimages into "feature subconcepts", with *k* corresponding to the number of convolutional filters. Each cluster was collapsed into its average, and one-versus-all SVMs were computed for each, which generated the convolutional filters. The outputs of this layer were passed through a max-pooling layer. Another set of convolutional filters and a max-pooling layer were performed on the outputs of the first max-pooled convolutional layer. The outputs of the second max-pooling layer were then fed into the ENN learning algorithm.

For all our convolutional layers we used SVM multipliers of 2, stride rate of 1x1 pixel, and max-pooling with non-overlapping 2x2 pixel boxes. The MNIST input images were padded out to be 32x32 pixels consistent with LeNet-5[8]. The smaller of the two cENNs in Table 1 was designed to be of similar dimensions to LeNet-5—the tolerated margin fraction adjusted to achieve this—and its learned filters are visualized in Fig. 3. The larger cENN was designed without any hyperparameter search, instead using the same parameters as ENNs before and without any differentia neuron pruning.

We visualized the filters both by plotting their weights and by computing the weighted average of all windows in the test set which lie on either side of the filter's hyperplane. Taking the filter neuron's output $y_i$ for each window, the weight applied to each was $|y_i - 0.5|$. For the second set of convolutional filters, the same was done, but taking the full receptive field from the original image that pertained to each filter.

Deliberative ENNs (dENNs)

There were multiple ways we found that dynamic, post-training deliberation could be implemented, so we chose to highlight the simplest method. When a test sample is given to the network, if there are two output probabilities that are within a factor of given ratio (which was 2 in all cases except for the TSP, where it was 10), deliberation is allowed to occur on that sample. The network then can uniformly increase or decrease the bias values of its subconceptual neurons in order to attempt to find a result where the output probabilities are well separated. It chooses to increase the bias factor if none of the subconceptual neurons are firing over 0.5 and decreases otherwise. The biases can be changed uniformly because computing the SVMs scales them all so that the weighted distance to the hyperplane is the same.

Gradient-descent-trained networks (GDNs)

Each GDN was trained using the same architecture as its comparison ENN. They were each trained using Keras with the Adam optimizer with default parameters and a categorical cross entropy loss function. 10-fold cross-validation was used to find the optimal batch size and number of training epochs for all GDNs except the GDNs computed for Fig. 2, which instead used 5-fold cross-validation.

Image datasets

The classification tasks tested in this paper include the following previously used datasets. The rectangles dataset was synthetically generated similar to previous work[52], with each image being a 28x28 black image with a white rectangular oriented horizontally or vertically, though we used filled rectangles. The convex dataset was also synthetically generated as in previous work[52], each image containing a filled convex or non-convex shape. For both the rectangle and convex data sets there were 50,000 training images and 10,000 test images.

The Modified National Institute of Standards and Technology (MNIST) dataset[8] is a widely used image set consisting of 70,000 28x28 grayscale images of handwritten digits 0 through 9. 60,000 images are available for training and validation, and the remaining 10,000 images have already been reserved in a standardized test set.

For the orientation problem, training images were 28x28 black images with a one-pixel-wide stripe across the full length or height of the image, which meant there were 56 total training images. The diagonal line and box outline datasets were generated as follows. For each pair of possible heights and widths of non-square rectangles in the image, we generated no more than 50 unique rectangles with randomly placed bottom-left corners. This rectangle's outline was drawn to make the box outline datasets, and one of its two diagonals was chosen randomly to make the diagonal line dataset. The lines dataset was the subset of diagonal images in which the lines were perfectly horizontal or vertical (i.e. width of one pixel).

Performance metrics

All reported training times are the measured wall times—starting once the training data was loaded and ending with the end of storing all of the network's parameters—on UTSW's BioHPC computing cluster on a single non-GPU node without parallelization (since we did not parallelize our ENN code). Error rates are from the test sets, which were held out from training and hyperparameter optimization. In order to assess how the size of the training set affected performance (Fig. 2), we trained ENNs and GDNs on random subsamples of MNIST, each subsample with the same number of images from each class. For each subsample size we repeated this 5 times. We used the same ENN hyperparameters, with 60 subconcepts and without any pruning to maintain a consistent network size (Fig. 2c). For each ENN and subsample we also generated a GDN, using 5-fold cross-validation to find the optimal batch size and training epochs for each. During ENN training, the images that served as support vectors were tallied and reported as well.

Boolean logic problem

The full 2-input truth table for all 16 Boolean functions, as displayed in Fig. 7a, was encoded into 64 individual samples. Each sample vector contained 18 features, the first two encod-ing the function input, and then one-hot encoding the function index, i.e. all 16 remaining features were set to 0 except for a single feature set to 1, corresponding to the Boolean function index. We found that the function input features needed to be scaled by a factor of 2 in order to get consistent clustering into ENN subconcepts, so True inputs were encoded as 1 and False entries encoded as -1.

The logic circuit in Fig. 7d involves 4 pairs of switches that depend on the function being called. When the inequality indicated is satisfied, both switches in the pair move upward, and otherwise the switches move downward. The plots in Fig. 11a were generated by feeding in interpolated function inputs and reporting the output True probability. The plots in Fig. 11b were found by tracing around this interpolated unit square for all 16 Boolean functions and measuring the difference between the network output and the correct value of the nearest corner (i.e. binary inputs).

Traveling salesman problem (TSP)

The TSP features a salesman trying to find the shortest possible route that takes him through all cities on a map and return home. Both the training and test sets consisted of samples with 55 features, 45 corresponding to the upper half of the inter-city distance matrix for a 10-city map, and the remaining 10 serving as a one-hot encoding of the current city, scaled up by 10. The cities were located on a map on the unit square. Cities that have already been visited are denoted in the distance matrix as being a distance of 10 from all other cities. The training set consisted of 90 samples corresponding to the maps with only on unvisited city. In order to teach a generalizable rule, the correct city to visit next was located a distance of 0 from the current city.

ENN training only allowed each subconcept to use as inputs its associated differentiae, and the initial concept layer used connections of weight 10 between subconcepts and their specific concept neurons which had bias -5. The output neurons used a sigmoid activation function. After training each network was asked to find a route for the test set maps. The distance matrix was given to the networks, which picked the next city to visit. The distance matrix was then altered by setting all distances from the previous city as 10 and changing the one-hot encoding to the new city. The network outputs corresponding to cities already visited were masked to prevent the possibility of endless loops.

The test set consisted of 5000 maps with the 10 cities all placed randomly. To serve as a reference for a greedy algorithm, each map was put through the greedy nearest-neighbor algorithm (i.e. choose for the next city the closest unvisited city). The test error reported for the TSP is the average difference in the route length found by the neural network compared to the nearest-neighbor algorithm.

Binary decision tree (BDT) problem

The problem is to find a BDT of minimum depth (i.e. cost) that fully reproduces a truth table. The depth of the tree is defined as the average depth necessary to classify each entry of the truth table (examples in Fig. 7e). Both the training and test sets consisted of samples with 1024 features corresponding to the label associated with each value in the 10-input truth table, encoded as zeros and ones. The training set consisted of 20 samples corresponding to all possible BDTs with only a single branch node.

After training, each network was asked to build full trees on the test set. This was done by feeding the truth table to the network and taking its output as the first branch node. Going down each of the branches in turn, if all entries on the branch were labeled the same, a leaf was placed at the end with the corresponding label. If more branch nodes were necessary, the truth table was reformed by taking the half corresponding to its side of the split and copying onto the other half, such that the already-split features no longer needed to be split. This new truth table was put through the network again with masking of the output choices that had already been split in order to prevent the possibility of an infinite tree. This was done until all branches had terminated in leaves.

The test set consisted of 5000 truth tables corresponding to trees of much greater depth. For each test sample, we generated a random BDT by allowing each node to branch with probability 0.7 and not allowing branches beyond a depth of 7. If a node did not branch, then its leaf labels were assigned randomly. The BDT's truth table was found and used as the test sample. Non-unique truth tables were discarded. To serve as a reference for a greedy algorithm, each tree was put through the CART algorithm with Gini impurity as the splitting criterion, using scikit-learn's DecisionTreeClassifier. The test error reported is the average difference in the tree depth found by neural network compared to the greedy CART algorithm.

Choosing optimal GDN network sizes for generalization

For the orientation problem and the TSP, we trained GDNs of varying layer widths, performing a grid search by scaling from 0 to twice the width of each ENN layer. We generated 10 GDNs for the orientation problem and 5 for the TSP. We then chose the architecture with the optimally generalizing GDN.

Seeding GDNs with ENN weights

In order to demonstrate the rarity of finding a generalizable solution with GDNs, we took the generalizing ENN, perturbed its weights by a small amount, and then trained it as a GDN. The perturbation consisted of adding a normally distributed value to all weights and biases, with the standard deviation being a given fraction of the mean weight magnitude for each layer separately. The fraction is the reported amount of noise added.

Network lesions

Lesions were performed in the second layer (subconceptual neurons in ENNs). Neurons were deleted sequentially, and test accuracy was calculated individually for each class. The sequence of neuron deletions was decided by using an ordering determined by hierarchical linkage clustering on the neurons' outputs on the test set, with the idea that neurons with similar firing patterns are physically located more closely together.

Distances to decision boundaries

For each sample in the test set correctly predicted by both the ENN and GDN—about 96% of MNIST and 99% of the rectangles test sets—20 target locations were chosen for interpolation. This target was either a test image from a different class or white noise (i.e. random black and white pixels) that the networks classified differently than the test image. Interpolating between the sample and the target, the point at which the network changed its predicted class was found and the average pixel difference was calculated (which is proportional to the $L_1$ distance to the boundary). The distribution of these distances for each sample are reported.

Robustness to noise

The error rate in classifying the test set was computed with increasing amounts of noise. The noise level was defined as the standard deviation of the Gaussian noise added to the test set, and the classification error was then computed. This was repeated 20 times for each noise level.

Robustness to adversarial attacks

To generate adversarial images, we used the fast gradient sign method (FGSM)[11], which calculates the sign of the gradient of the loss function $L$ with respect to the inputs $\boldsymbol{x}$, $\text{sgn}(\nabla_{\boldsymbol{x}} L)$, and then scales this vector by a small $\varepsilon$. We increased the value of $\varepsilon$ until we found the minimum perturbation necessary to cause misclassification, $\varepsilon_{\min}$. For both the GDN and ENN we computed $\nabla_{\boldsymbol{x}} L$ for each image, with the loss function $L$ for both being the cross-categorical entropy function used to train the GDN. This network-specific perturbation was allowed to scale separately to find $\varepsilon_{\min}$ for each network.

# Appendix B: Pseudocode generated from ENNs

ENN pseudocode for orientation problem

The algorithm examines each pixel in the image and determines whether that pixel's column or row has a greater number of white pixels. It then looks at each row and column to see if its constituent pixels favored the column or row designation. The output is whether there are more rows or columns that appear to be populated with white pixels in the image.

```
ENN_Orientation(n,I):
// n is the square image dimension (28 in our experiments)
// I is the image

// Each differentia represents a pixel in I, whether its column or row has more white
D = zeros(n,n)
for x = 0 to n-1:
   for y = 0 to n-1:
      D[x,y] = sign(sum(I[:,n]) - sum(I[n,:]))

// For each row and column, see if it can be said to be present
S = zeros(n,2)
for row = 0 to n-1:
   S[row,0] = sign(sum(D[row,:]) – (n-1)*(sum(d)-n))
for col = 0 to n-1:
   S[col,1] = sign(sum(D[:,col]) – (n-1)*(sum(d)-n))

// Return horizontal if there are more rows found, and columns otherwise
if sum(S[:,0])>sum(S[:,1]):
   return HORIZONTAL
else:
   return VERTICAL
```

ENN pseudocode for traveling salesman problem (TSP)

The algorithm essentially performs the nearest-neighbor algorithm. It looks at each city transition and determines which ones actually start from the current city and then which new city wins the most comparisons to other possible cities.

```
dENN_TSP(n,M,c)
// n is the number of cities in the map (10 in our experiments)
// M is a matrix of distances between cities (0 is the current city)
// c is the index of the current city

// Most differentiae represent the comparison of whether the distance
// from city 1 to city 2 is shorter than city 1 to city 3.
// The other differentiae compare the marker for the city 1.
D1 = zeros(n,n,n)
D2 = 0.5*ones(n,n)
for c1 = 0 to n-1:
   for c2 = 0 to n-1 but not c2:
      for c3 = 0 to n-1 but not (c1 or c2):
         D1[c1,c2,c3] = sign(M[c1,c3]-M[c1,c2])
      if c1 == c:
         D2[c1,c2] = 1
      if c2 == c:
         D2[c1,c2] = 0

// Subconcept represent the transition of one city to a different city
S = zeros(n,n)
for c1 = 0 to n-1:
   for c2 = 0 to n-1:
      if c1 == c2:
         continue
      S[c1,c2] = D2[c1,c2] + sum(D1[c1,c2,:])

// Search for which city-to-city transition won the most comparisons
max_s = 0
last_change = 0
while true:
   concepts = arg(S>max_s)
   if length(concepts) == 1:
      break
   else if length(concepts) == 0:
      max_s -= 1
      last_change = -1
   else:
      if last_change < 0:
         break
      max_s += 1
      last_change = 1

// Return city 2 from the winning subconcept
return concepts[0,1]
```

ENN pseudocode for binary decision tree (BDT) problem

The algorithm looks at each feature and tries to see how many of the truth table entries are True or False for either value of that feature. It makes these comparisons for each feature for all truth table assignments, and then picks the one that has the best gain.

```
dENN_BDT(n,T,L):
// n is the number of features in the truth table (10 here)
// T is the truth table (n x 2^n)
// L is the label vector (1 x 2^n)

// Initialize differentiae, which looks at pairs of truth table values
D_00 = zeros(n,n)
D_01 = zeros(n,n)
D_10 = zeros(n,n)
D_11 = zeros(n,n)

// For each feature pair, determine which split leads to more lopsided truth tables
for f1 = 0 to n-1:
   for f2 = 0 to n-1:
      f_counts = zeros(2,2)
      for v1 = 0 to 1:
         for v2 = 0 to 1:
            f_counts[v1,v2] = sum(L[t] for t=0 to 2^n if T[t,[f1,f2]] == [v1,v2])
      D_11[f1,f2] = sign(f_counts[1,0] - f_counts[0,1])
      D_10[f1,f2] = sign(f_counts[1,1] - f_counts[0,0])
      D_01[f1,f2] = sign(f_counts[0,0] - f_counts[1,1])
      D_00[f1,f2] = sign(f_counts[0,1] - f_counts[1,0])

// Sum how many comparisons each feature has won
S = zeros(n,2)
for f = 0 to n-1:
   S[f,0] = sum(D_00[f,:]) + sum(D_01[f,:])
   S[f,1] = sum(D_10[f,:]) + sum(D_11[f,:])

// Iteratively search for which feature had the most wins
max_s = n
last_change = 0
while true:
   concepts = arg(S>max_s)
   if length(concepts) == 1:
      break
   else if length(concepts) == 0:
      max_s -= 1
      last_change = -1
   else:
      if last_change < 0:
         break
      max_s += 1
      last_change = 1

// Return only one of the possible winners
return concepts[0,0]
```